\documentclass[runningheads]{llncs}

\usepackage[T1]{fontenc}

\usepackage{graphicx,verbatim}
\usepackage{tabularx}
\usepackage{subcaption} 
\usepackage{makecell}
\usepackage{amsmath}
\usepackage{float}
\usepackage{amsmath}   
\usepackage{amsfonts} 
\usepackage{hyperref}
\usepackage{xurl}
\usepackage{color}

\usepackage[table]{xcolor}
\definecolor{gold}{RGB}{250,246,189}
\definecolor{silver}{RGB}{196,229,247}
\definecolor{bronze}{RGB}{230,191,152}

\begin{document}

\title{One Model to Magnify Them All: Efficient Scale-Invariant Histopathology via Conditional Normalization and Continuous Magnification Training}
\titlerunning{One Model to Magnify Them All: Efficient Scale‑Invariant Histopathology}
 
\author{Agnieszka Florkowska \inst{1, 2}\orcidID{0009-0005-2544-5510} \and
Henning Müller\inst{3,4}\and
Marek Wodzinski\inst{1, 2}\orcidID{0000-0002-8076-6246}}
\authorrunning{A. Florkowska et al.}

\institute{
Sano Centre for Computational Medicine, Krakow, Poland \and  
AGH University of Krakow, al. Adama Mickiewicza 30, 30-059 Kraków, Poland \email{aflorkowska@agh.edu.pl} \and
Institute of Informatics, HES-SO
Valais-Wallis, Sierre, Switzerland \and
Faculty of Medicine, University of Geneva (UNIGE), Geneva,
Switzerland
}

\maketitle              

\begin{abstract}

Whole slide images (WSIs) in digital histopathology are acquired at discrete magnification levels encoding complementary diagnostic information from global tissue architecture to fine-grained cellular morphology. Yet, deep learning models remain sensitive to scale variation. Existing magnification-invariant methods rely on multi-scale architectures at predefined discrete resolutions, while in clinical deployment the acquisition magnification
varies continuously,
rarely aligns with a model's fixed training resolution, and intermediate
scales are common, so robust coverage otherwise demands a costly ensemble of magnification-specific models. We propose Conditional Layer Normalization (CLN), a lightweight mechanism that generates affine normalization parameters from input pixel size via a small MLP, integrated into standard CNN architectures for both WSI classification and segmentation. Trained on patches sampled continuously across a range of pixel sizes, the model decouples inference from scanner-dependent magnification and generalizes to arbitrary, previously unseen scales at test time. On the PANDA prostate cancer dataset, our approach on average matches or exceeds independently trained single-magnification models and ranks among
the top three performers at every evaluated magnification, including those
unseen during training. This collapses a five-model ensemble into a single network and reduces training, and inference cost roughly $4$-$5\times$,
while leaving the multiply-accumulate count unchanged.
The code is available at: \url{https://github.com/aflorkowska/OneModelToMagnifyThemAll}.

\end{abstract}

\section{Introduction}

Histopathological diagnosis inherently requires reasoning across
spatial scales: low-magnification views reveal tissue architecture and glandular organization, while high-magnification fields expose cellular morphology and nuclear atypia. The dominant strategies for handling this multi-scale nature achieve robustness at a steep efficiency cost: training, storage, and inference all scale with the number of magnification levels. In clinical settings where computational budget and deployment constraints are limiting, this multiplication of cost is a practical barrier to deployment. Most architectures sidestep the problem by training and evaluating at a single fixed magnification~\cite{srinidhi2021deep}, trading cross-scale robustness for simplicity.

Existing approaches address scale variation through several 
paradigms. Multi-scale CNN architectures process patches at 
multiple magnifications, dedicating separate computational streams, either via late feature fusion~\cite{tokunaga2019adaptive}, dedicated branches per 
magnification level~\cite{marini2021multi}, or multi-scale 
multiple instance learning frameworks~\cite{hashimoto2020multi}. 
Hierarchical transformers such as Hierarchical Image Pyramid Transformer (HIPT)~\cite{chen2022scaling} 
aggregate spatial tokens across nested pixel regions, yet operate 
within a single fixed magnification level. Foundation models such 
as Virchow2~\cite{zimmermann2024virchow2} incorporate multiple 
discrete magnification levels during pretraining ($5\times$, 
$10\times$, $20\times$, $40\times$), but remain bound to those 
predefined scales without explicit conditioning on acquisition 
resolution. Self-supervised methods such as Magnification-Aware Distillation (MAD)~\cite{gokmen2025magnification} 
enforce cross-scale consistency by linking low- and high-magnification 
representations via distillation, but target representation learning 
rather than task-specific inference at arbitrary magnifications.
Despite their diversity, these methods share three fundamental 
limitations:
(i) \textit{Cross-magnification domain shift}: models trained at a 
fixed magnification suffer performance degradation when evaluated 
at unseen scales~\cite{tokunaga2019adaptive,marini2021multi}, as 
visual statistics differ dramatically across resolutions;
(ii) \textit{Computational overhead}: covering the full range of
acquisition magnifications at deployment forces an ensemble of
magnification-specific models, multiplying training, and inference cost
with the number of magnification levels~\cite{hashimoto2020multi};
(iii) \textit{Discrete scale constraints}: existing methods - 
whether single- or multi-magnification - operate exclusively at 
scales seen during training, providing no mechanism to handle 
arbitrary or previously unseen intermediate 
resolutions~\cite{gokmen2025magnification}.

 \begin{figure}[htbp]
 \centering
\includegraphics[width=\textwidth]{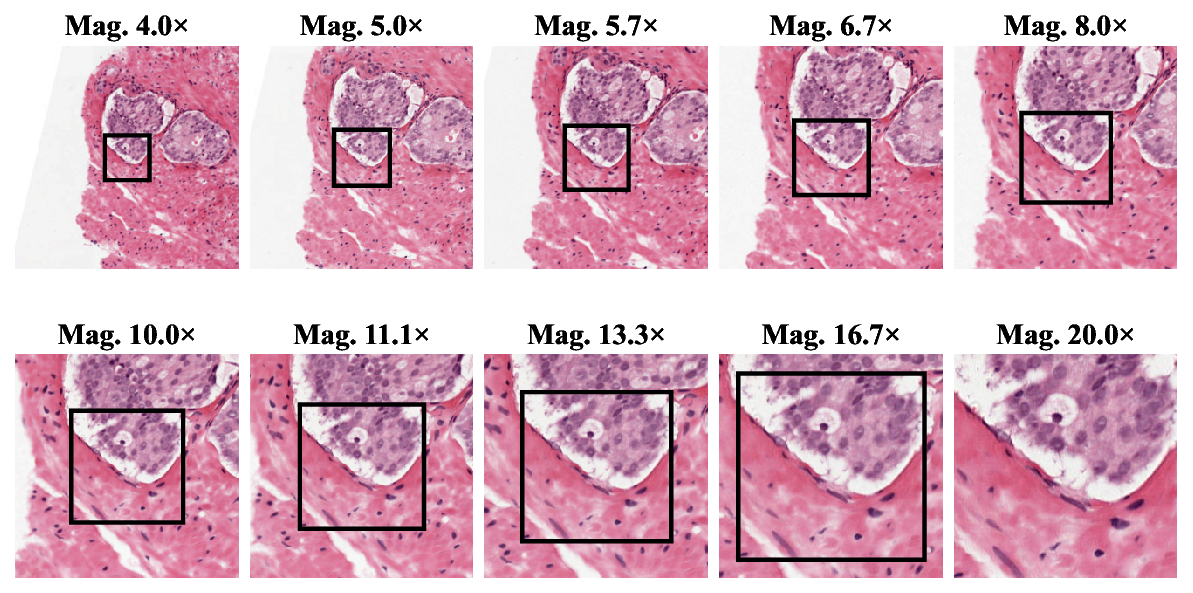}
\caption{Multi-scale histopathological view of a prostate biopsy specimen at magnifications ranging from 20$\times$ to 4$\times$, illustrating progressive transition from cellular detail to overall glandular architecture (PANDA dataset \cite{litjens2020prostate}).}
\label{fig:zooming}
\end{figure}

\textbf{Contribution: } We address these limitations by proposing an efficient and scalable Conditional Layer Normalization (CLN), a normalization layer that dynamically modulates internal feature representations as a function of input pixel size. We integrate CLN into standard ResNet \cite{he2016deep} and U-Net \cite{ronneberger2015u} architectures for classification and segmentation, respectively, without strongly modifying the network itself. Critically, models are trained on patches sampled continuously across a range of pixel sizes, exposing the network to a continuum of resolutions rather than a discrete set. This continuous sampling strategy is the key departure from prior work: it enables robust generalization to arbitrary, previously unseen magnification levels at inference time without requiring multi-scale input representations or model ensembles. Experiments on the PANDA dataset demonstrate that a single CLN-conditioned model on average matches or exceeds independently trained single-magnification baselines across both classification and segmentation, remaining within the top three models at every tested
magnification without requiring retraining or knowledge of the acquisition magnification at deployment. Single-magnification baselines, by contrast, only rank highest near their own training magnification and degrade substantially elsewhere. This approach collapses a five-model ensemble into
one network and reduces training and inference cost roughly $4$-$5\times$ with no increase in multiply-accumulate operations.

\section{Methods}

\subsection{Conditional Layer Normalization}

\begin{figure}[H]
    \centering
   \includegraphics[width=\textwidth]{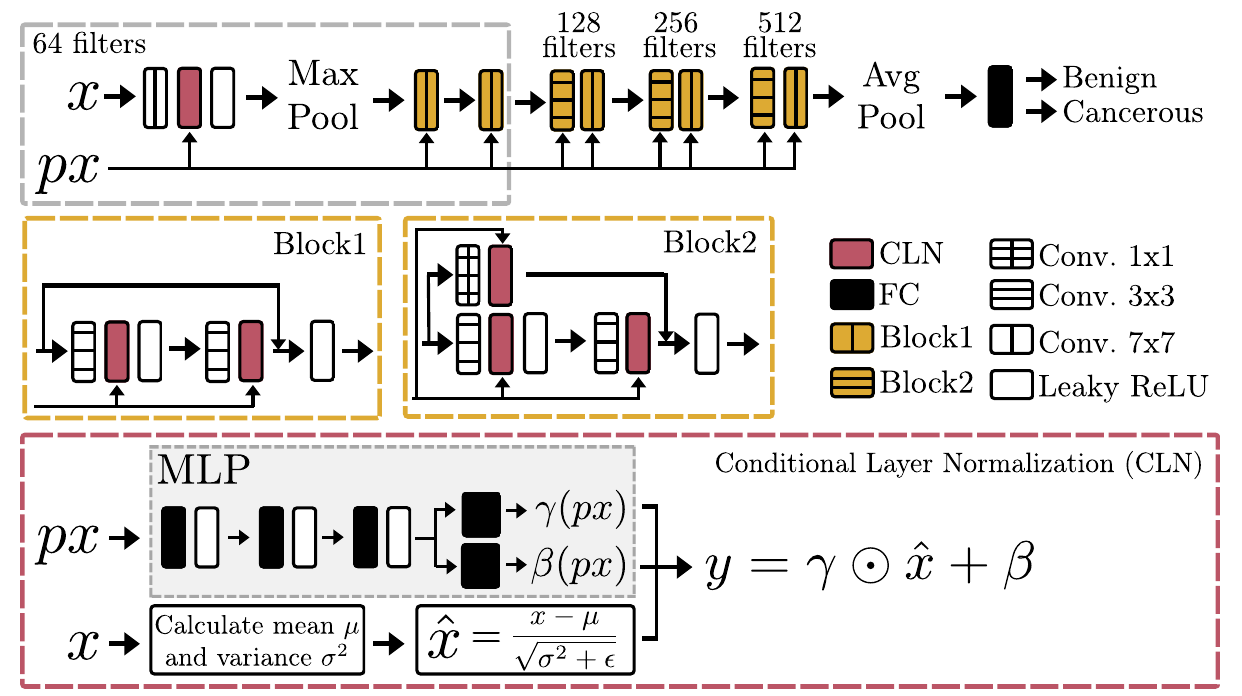}
    \caption{Conditional Layer Normalization (CLN) architecture and its integration into ResNet. The MLP maps input pixel size $px \in \mathbb{R}^{B \times 2}$ to affine parameters $\gamma(px)$ and $\beta(px)$, which modulate normalized feature maps $\hat{x}$ computed from patch input $x \in \mathbb{R}^{B \times \ldots \times C}$,  producing the conditioned output $y = \gamma(px) \odot \hat{x} + 
    \beta(px)$. In U-Net, CLN is applied analogously at each encoder and decoder stage.}
    \label{fig:conditionalBlock}
\end{figure}

Standard layer normalization applies fixed learned affine parameters $\gamma$ and $\beta$ to normalized features, independent of the acquisition context. We replace this with \textbf{Conditional Layer Normalization (CLN)}, which generates scale $\gamma$ and shift $\beta$ parameters dynamically from the input pixel size $px \in \mathbb{R}^2$ via a lightweight multilayer perceptron (MLP), as shown in Figure~\ref{fig:conditionalBlock} and Equation~\ref{eq:norm}:

\begin{equation}
\text{CLN}(x; px) = \gamma(px) \odot \hat{x} + \beta(px), \quad 
\hat{x} = \frac{x - \mu}{\sqrt{\sigma^2 + \epsilon}},
\label{eq:norm}
\end{equation}

where $x$ denotes the input feature map, $\mu$ and $\sigma^2$ are 
its mean and variance, $\epsilon = 1\times 10^{-5}$ is a numerical 
stability constant, and $\gamma(px)$, $\beta(px)$ are scale and shift 
parameters dynamically predicted by independent three-layer MLPs 
with LeakyReLU activations, each conditioned on pixel size $px$.

Pixel size, a physical hardware-independent quantity expressed in micrometers per pixel, is used as the conditioning signal rather than the scanner-dependent magnification, ensuring generalization across multiple magnifications.

The hidden dimensionality of the conditioning MLP at each network stage matches the feature dimensionality of the corresponding normalization layer, allowing conditioning capacity to scale naturally with representational complexity. The output projection layers are initialized to produce $\gamma = 1$ and $\beta = 0$, so CLN reduces to standard layer normalization at initialization and deviates only as scale-dependent adaptation emerges through training. No other architectural modifications are required, making CLN a drop-in replacement compatible with any normalization-equipped architecture. As illustrated in Figure~\ref{fig:conditionalBlock}, CLN is applied at every resolution stage of both the ResNet encoder (for classification) and the U-Net encoder-decoder (for segmentation).

\subsection{Continuous Magnification Training}

Rather than training at a discrete set of magnification levels, we sample patch pixel sizes uniformly from a continuous range at each training iteration. This exposes the model to a dense continuum of resolutions, preventing overfitting to scanner-specific discrete scales and enabling generalization to intermediate magnifications not encountered during training. At inference, the model accepts any pixel size, decoupling deployment from hardware acquisition settings. This continuous training strategy, combined with CLN conditioning, constitutes the core contribution of this work.

\subsection{Experimental setup}

\noindent\textbf{Architectures.} We adopt ResNet (detailed in Figure~\ref{fig:conditionalBlock}) for binary patch-level classification and U-Net (32 initial filters) for pixel-level segmentation. These architectures provide a transparent and reproducible benchmark for isolating the contribution of CLN from confounding architectural factors.

\noindent\textbf{Training.} All models were trained with no prior knowledge under identical hyperparameters. We used the AdamW optimizer ($\text{lr}=0.001$) with cosine annealing learning rate scheduling ($T_{\max}=200$, $\eta_{\min}=10^{-5}$) and early stopping ($\text{patience}=100$, $\text{min delta}=10^{-4}$) monitored on validation loss. Batch size was 128; 10,000 patches were sampled per epoch and resized to $224\times224$ pixels.

\noindent\textbf{Models.} Five single-magnification baselines were trained independently at $20\times$, $13.3\times$, $10\times$, $6.7\times$, $5\times$. A single CLN-conditioned model was trained continuously across the same range.

\noindent\textbf{Evaluation.} All models were evaluated at both seen ($20\times$, $13.3\times$, $10\times$, $6.7\times$, $5\times$) and unseen intermediate magnifications ($16.7\times$, $11.1\times$, $8\times$, $5.7\times$, $4.0\times$), assessing generalization to unencountered resolutions. Inference was performed without patch overlap. Figure~\ref{fig:zooming} illustrates the visual transition across scales, from fine cellular detail at high magnification to overall glandular architecture at low magnification. Statistical comparisons were conducted using the Wilcoxon signed-rank test: Dice score distributions per WSI for segmentation, and predicted probability error for classification.

\subsection{Dataset}

We used the PANDA dataset \cite{litjens2020prostate}, which provides prostate biopsy WSIs with strong pixel-level Gleason grade annotations, allowing evaluation on both classification and segmentation tasks. The highest resolution available in the public release of the dataset corresponds to 20× magnification. Background regions were excluded using tissue masks \cite{jMultistain2024,j2024improving}. The original six-class annotation schema was remapped to three classes: stroma (class 0), benign epithelium (class 1), and cancerous epithelium (Gleason grades 3–5 merged; class 2). For classification, patch-level labels were derived from the dominant class, yielding a binary benign vs. malignant task. Slides with incomplete annotations or insufficient class representation were excluded. The remaining slides were stratified by class combination and partitioned at the slide level - ensuring no data leakage across splits - into training, validation, and test sets ($\min 20\%\ \text{of total samples},\ 600$) slides for test, $10\%$ of the remainder for validation. The number of patches extracted per slide was determined dynamically.

A custom data loader supports patch extraction at either a fixed or continuously sampled pixel size drawn uniformly from the training range. Class imbalance was mitigated through the use of a weighted random sampling strategy during model training. The corresponding implementation is provided in the associated code repository. Augmentation consisted of random horizontal and vertical flips ($p=0.5$), random rotation, and z-normalization per-patch.

\section{Results}

\begin{figure}[H]
    \centering
    \begin{subfigure}[b]{0.45\textwidth}
        \includegraphics[width=\textwidth]{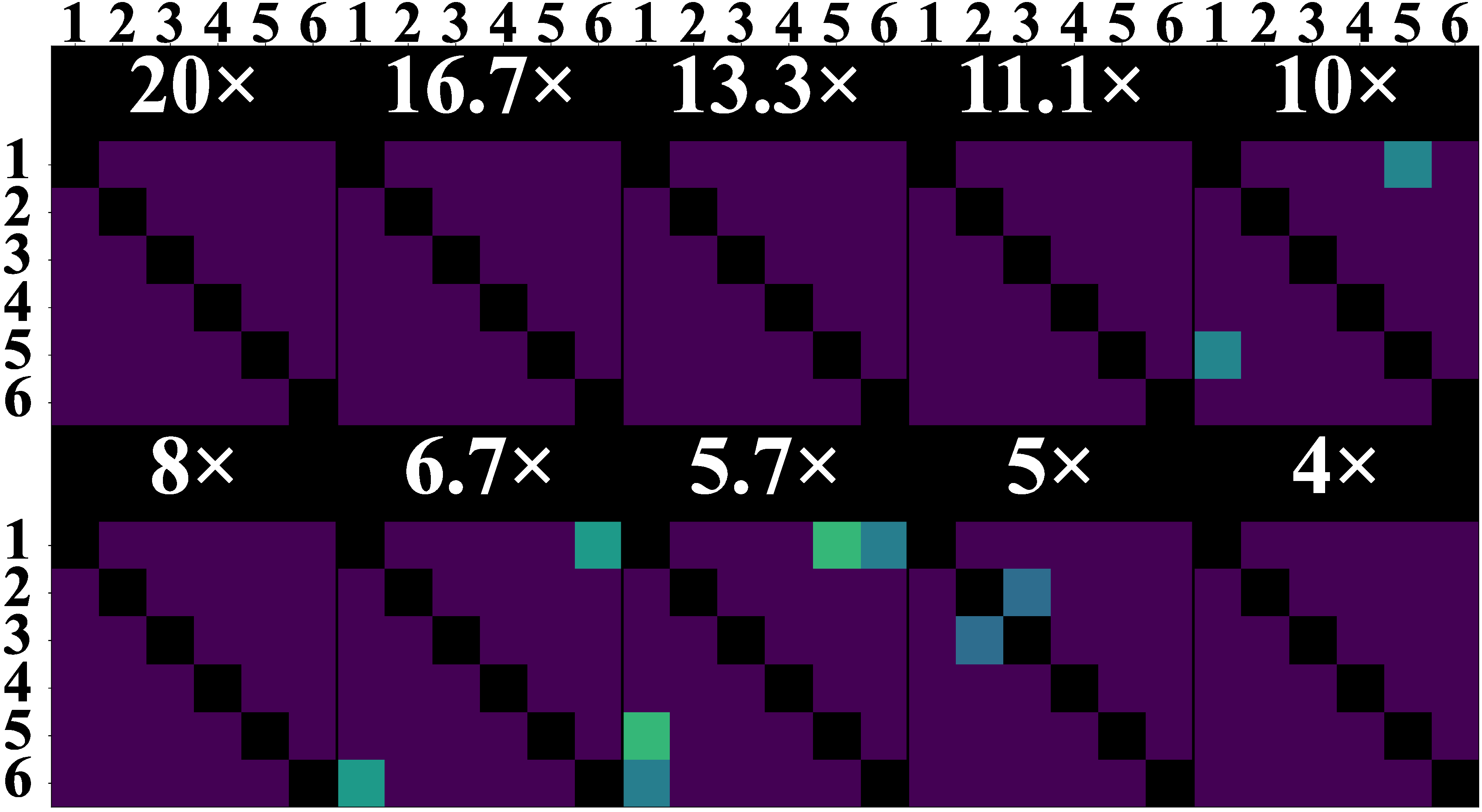}
        \caption{ResNet classification}
        \label{fig:classification}
    \end{subfigure}
    \hfill
    \begin{subfigure}[b]{0.45\textwidth}
        \includegraphics[width=\textwidth]{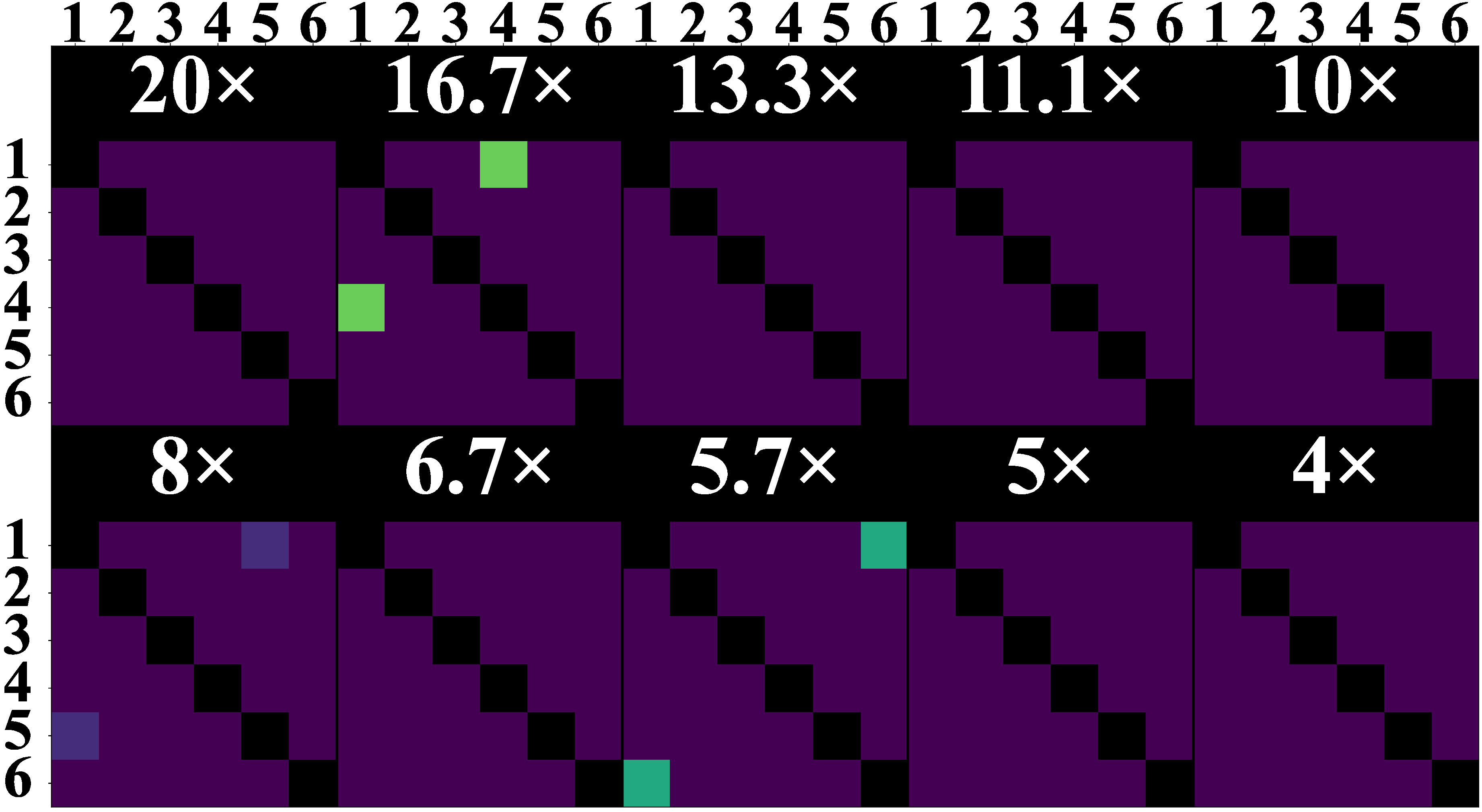}
        \caption{UNet segmentation}
        \label{fig:segmentation}
    \end{subfigure}
    \begin{minipage}[b]{\textwidth}
        \centering
        \includegraphics[width=0.7\textwidth]{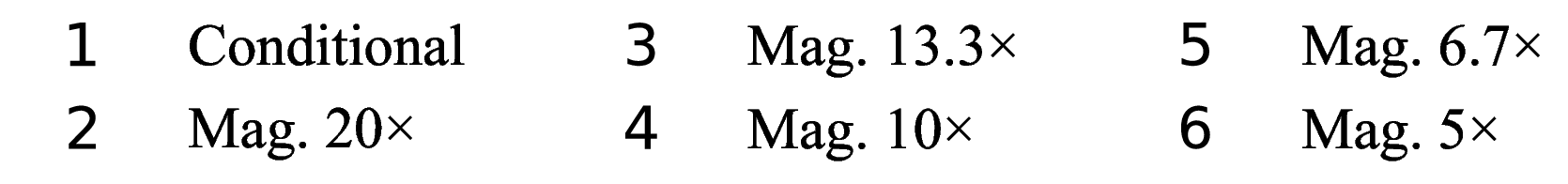}
    \end{minipage}
     \begin{minipage}[b]{\textwidth}
        \centering
        \includegraphics[width=0.7\textwidth]{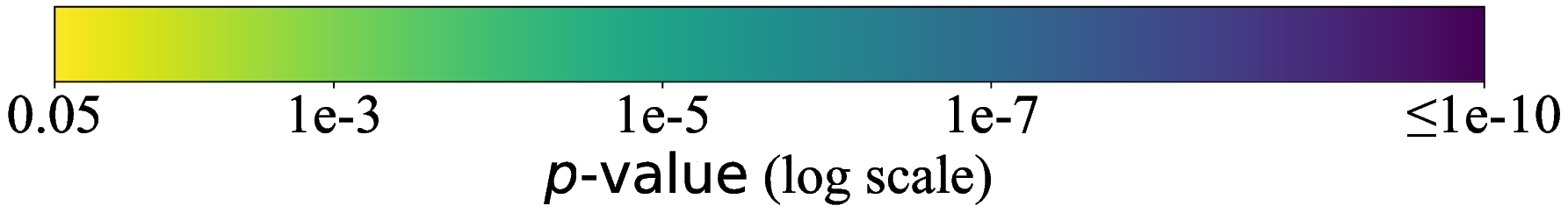}
    \end{minipage}
    \caption{Pairwise statistical comparison (Wilcoxon signed-rank test) between the CLN-conditioned model and single-magnification baselines across all evaluated magnifications for (a) classification and (b) segmentation. Color encodes the $p$-value on a logarithmic scale, ranging from $0.05$ (least significant) to $\leq 10^{-10}$ (most significant. All off-diagonal pairs are statistically significant ($p < 0.05$). The majority of pairs extend far below the color scale upper bound: 88.7\% have $p < 10^{-200}$ for classification and 96.7\% for segmentation.}
    \label{fig:p_value}
\end{figure}

\begin{figure}[H]
    \centering    
    \begin{subfigure}[b]{0.48\textwidth}
        \centering
        \includegraphics[width=0.85\textwidth]{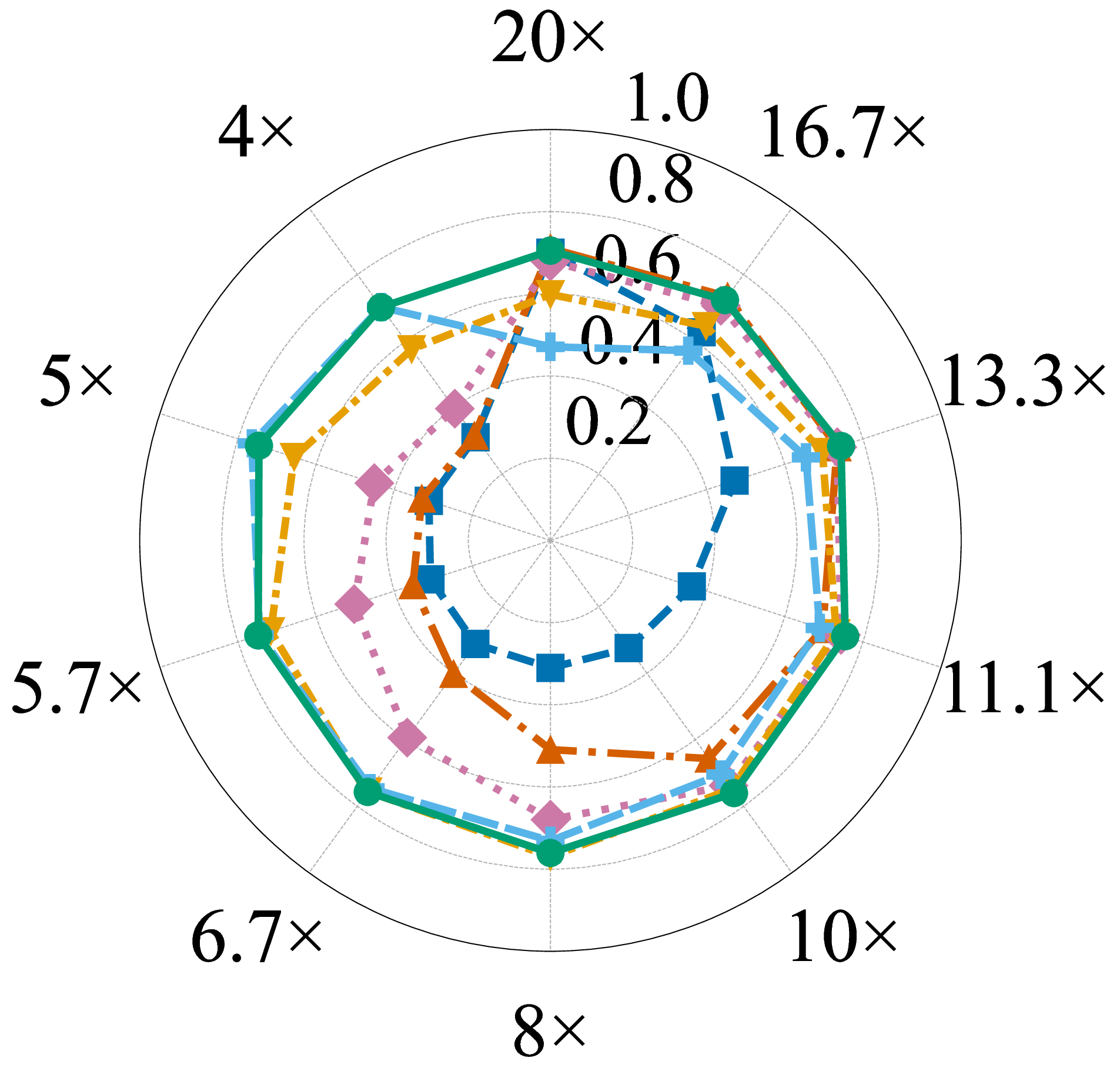}
        \caption{ResNet classification - F1 score}
        \label{fig:RP_classification}
    \end{subfigure}
    \hfill
    \begin{subfigure}[b]{0.48\textwidth}
        \centering
        \includegraphics[width=0.85\textwidth]{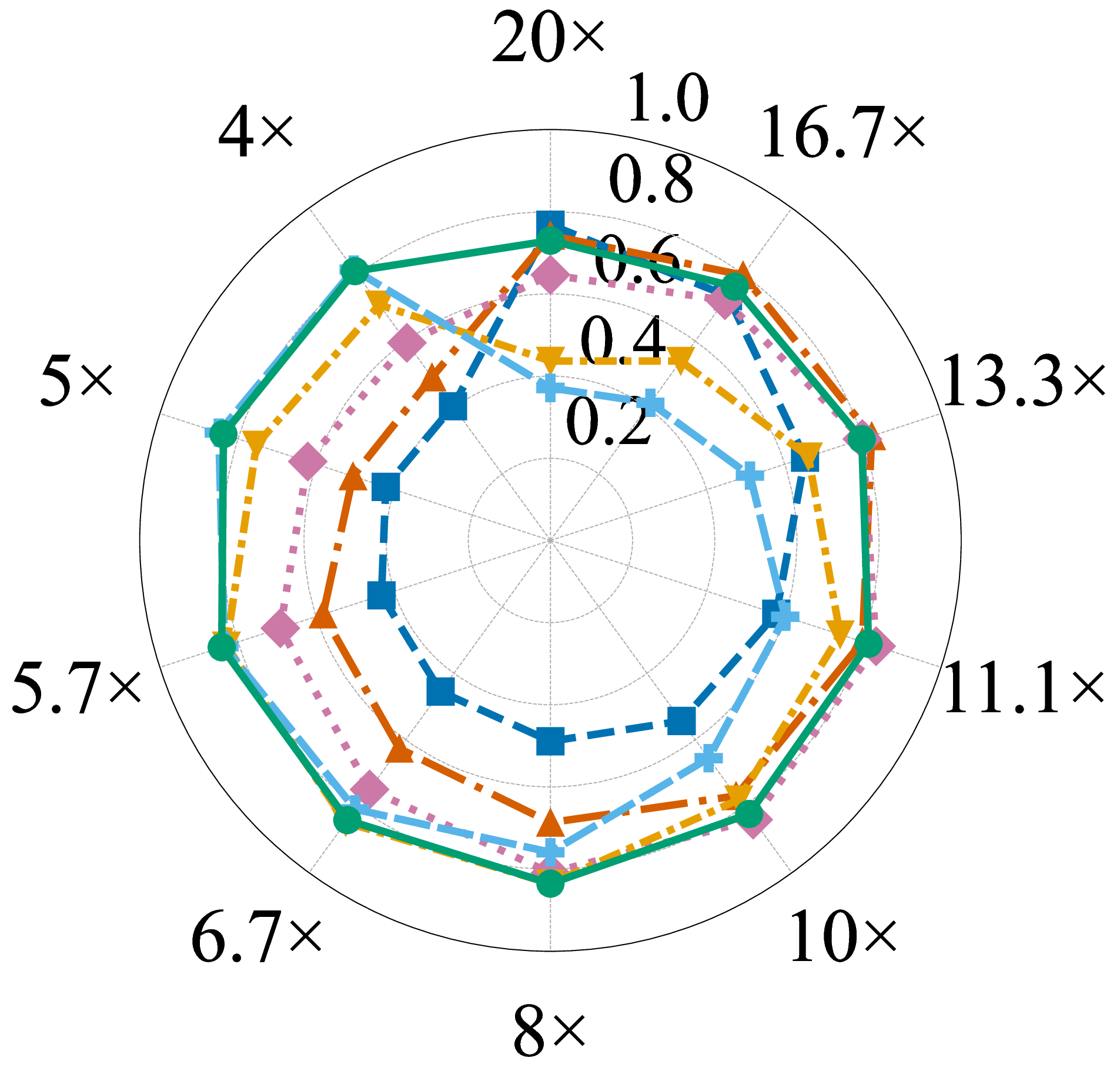}
        \caption{UNet segmentation - Dice score}
        \label{fig:RP_segmentation}
    \end{subfigure}
      \begin{minipage}[b]{\textwidth}
        \centering
        \includegraphics[width=0.8\textwidth]{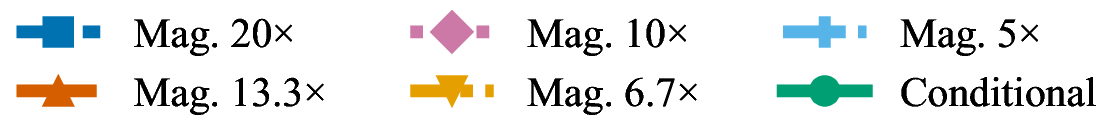}
    \end{minipage}
  \caption{Model performance with respect to the magnification level in the PANDA prostate biopsies, for both downstream tasks. Each plot compares models trained at specific magnifications and evaluated across multiple inference magnifications.}
    \label{fig:radial_plots}
\end{figure}

Table~\ref{tab:f1_dice} reports F1 (classification) and Dice 
(segmentation) scores across all evaluated magnifications. Figure~\ref{fig:radial_plots} visualizes the corresponding performance profiles for both tasks. The CLN-conditioned model exhibits consistent, stable performance throughout the evaluated range, including at unseen intermediate scales, while single-magnification models degrade predictably when evaluated outside their training resolution. Statistical analysis (Fig. \ref{fig:p_value}) confirms that observed differences between the conditioned model and single-magnification baselines are statistically significant, ruling out chance variation.

Per-batch compute was profiled with CUDA events at batch size 128, matching the tiled WSI regime in computational pathology.
Inference cost is operationally decisive: training is one-time, inference recurs on every slide. Five separate ResNets sum to a median of $57.90$~ms per batch against $14.05$~ms for CLN ($-75.7\%$), and the five U-Nets to $281.95$~ms against $56.79$~ms ($-79.9\%$). Training shows the same pattern per step, with the five ResNets at $269.07$~ms against $67.89$~ms for CLN ($-74.8\%$) and the five U-Nets at $1148.26$~ms against $230.71$~ms ($-79.9\%$). CLN adds $+13.4\%$/$+45.1\%$ parameters to ResNet/U-Net yet leaves the multiply-accumulate count unchanged ($1.81$/$10.48$~GMAC), as the conditioning MLPs act only on the two-element pixel-size vector.

\section{Discussion and Conclusion}

These results validate CLN as a viable alternative to per-resolution model
ensembles, shifting the paradigm toward a unified, metadata-aware architecture. Such an ensemble is the appropriate deployment baseline: acquisition magnification varies continuously
across incoming slides, and diagnostically relevant scales often fall between scanner presets. These intermediate magnifications are not natively handled by any single-magnification model, yet CLN covers them with a single network. While this work constitutes an initial validation of the concept on a
single dataset and CNN-based backbones, the consistency of results across
both seen and unseen magnifications suggests the approach is promising,
though broader validation is needed to establish robustness beyond this
setting.

\begin{table}[!ht]
    \centering
    \caption{Single-magnification models (rows) are evaluated at all magnification levels (columns) to assess cross-magnification generalization. The conditional model is trained to natively handle variable magnification. The top three values in each column are highlighted: yellow (1st, bold), blue (2nd), and brown (3rd). Note that the stability of CLN is maintained despite changes in pixel size.}
    \label{tab:f1_dice}
    \begin{subtable}[t]{\textwidth}
        \centering
        \caption{ResNet classification - F1 score}
        \begin{tabularx}{\textwidth}{|m{2cm}|*{10}{>{\centering\arraybackslash}X|}}
        \hline
        \multicolumn{1}{|c|}{} & \multicolumn{10}{c|}{\textbf{Magnification used during inference (evaluation)}} \\ \hline
        \textbf{Model (training mag.)} & \textbf{20$\times$} & \textbf{16.7$\times$} & \textbf{13.3$\times$} & \textbf{11.1$\times$} & \textbf{10$\times$} & \textbf{8$\times$} & \textbf{6.7$\times$} & \textbf{5.7$\times$} & \textbf{5$\times$} & \textbf{4$\times$} \\ \hline
        20$\times$ & \cellcolor{bronze}0.700 & 0.626 & 0.471 & 0.362 & 0.324 & 0.310 & 0.311 & 0.307 & 0.311 & 0.310 \\ 
        
        13.3$\times$ & \textbf{\cellcolor{gold}0.711} & \textbf{\cellcolor{gold}0.733} & \cellcolor{silver}0.734 & 0.697 & 0.656 & 0.509 & 0.401 & 0.352 & 0.329 & 0.315 \\ 
        
        10$\times$ & 0.679 & \cellcolor{bronze}0.706 & \cellcolor{silver}0.734 & \cellcolor{bronze}0.744 & \cellcolor{bronze}0.738 & 0.680 & 0.593 & 0.502 & 0.451 & 0.396 \\ 
        
        6.7$\times$ & 0.599 & 0.645 & \cellcolor{bronze}0.696 & \cellcolor{silver}0.732 & \cellcolor{silver}0.749 & \textbf{\cellcolor{gold}0.770} & \cellcolor{silver}0.756 & \cellcolor{bronze}0.716 & \cellcolor{bronze}0.656 & \cellcolor{bronze}0.575 \\ 
        
        5$\times$ & 0.471 & 0.571 & 0.654 & 0.689 & 0.704 & \cellcolor{bronze}0.731 & \cellcolor{bronze}0.746 & \cellcolor{silver}0.745 & \textbf{\cellcolor{gold}0.765} & \textbf{\cellcolor{gold}0.702} \\ 
        
        Conditional & \cellcolor{silver}0.705 & \cellcolor{silver}0.723 & \textbf{\cellcolor{gold}0.744} & \textbf{\cellcolor{gold}0.755} & \textbf{\cellcolor{gold}0.760} & \cellcolor{silver}0.761 & \textbf{\cellcolor{gold}0.757} & \textbf{\cellcolor{gold}0.748} & \cellcolor{silver}0.746 & \cellcolor{silver}0.701 \\ 
        \hline
        \end{tabularx}
    \end{subtable}
    \begin{subtable}[t]{\textwidth}
    \centering
    \caption{UNet segmentation - Dice score}
    \begin{tabularx}{\textwidth}{|m{2cm}|*{10}{>{\centering\arraybackslash}X|}}
    \hline
    \multicolumn{1}{|c|}{} & \multicolumn{10}{c|}{\textbf{Magnification used during inference (evaluation)}} \\ \hline
    \textbf{Model (training mag.)} & \textbf{20$\times$} & \textbf{16.7$\times$} & \textbf{13.3$\times$} & \textbf{11.1$\times$} & \textbf{10$\times$} & \textbf{8$\times$} & \textbf{6.7$\times$} & \textbf{5.7$\times$} & \textbf{5$\times$} & \textbf{4$\times$} \\ \hline
    20$\times$ & \textbf{\cellcolor{gold}0.768} & \cellcolor{bronze}0.735 & 0.652 & 0.578 & 0.543 & 0.489 & 0.455 & 0.433 & 0.422 & 0.404 \\ 
    13.3$\times$ & \cellcolor{silver}0.742 & \textbf{\cellcolor{gold}0.798} & \textbf{\cellcolor{gold}0.822} & \cellcolor{bronze}0.799 & 0.769 & 0.687 & 0.626 & 0.582 & 0.506 & 0.491 \\ 
    10$\times$ & 0.646 & 0.723 & \cellcolor{silver}0.798 & \textbf{\cellcolor{gold}0.834} & \textbf{\cellcolor{gold}0.840} & \cellcolor{bronze}0.809 & 0.749 & 0.692 & 0.621 & 0.594 \\ 
    6.7$\times$ & 0.437 & 0.540 & 0.660 & 0.742 & \cellcolor{bronze}0.780 & \textbf{\cellcolor{gold}0.835} & \textbf{\cellcolor{gold}0.848} & \cellcolor{bronze}0.830 & \cellcolor{bronze}0.753 & \cellcolor{bronze}0.709 \\ 
    5$\times$ & 0.371 & 0.414 & 0.511 & 0.601 & 0.655 & \cellcolor{bronze}0.759 & \cellcolor{bronze}0.809 & \cellcolor{silver}0.835 & \textbf{\cellcolor{gold}0.849} & \textbf{\cellcolor{gold}0.815} \\ 
    Conditional & \cellcolor{bronze}0.730 & \cellcolor{silver}0.763 & \cellcolor{bronze}0.797 & \cellcolor{silver}0.814 & \cellcolor{silver}0.823 & \textbf{\cellcolor{gold}0.835} & \cellcolor{silver}0.841 & \textbf{\cellcolor{gold}0.842} & \cellcolor{silver}0.837 & \cellcolor{silver}0.810 \\ 
    \hline
    \end{tabularx}
\end{subtable}
\end{table}

We demonstrated that a single model conditioned on input pixel size via Conditional Layer Normalization on average matches or exceeds independently trained single-magnification models on both WSI
classification and segmentation, remaining among the top three models at every evaluated magnification, including those unseen during training. The continuous magnification training strategy generalizes to unseen intermediate scales without multi-scale input representations, multi-model ensembles, or task-specific architectural modifications. By replacing five independent models with a single conditioned network,
the approach yields a roughly $4$-$5\times$ reduction in training and inference cost at zero per-input FLOP penalty, with direct implications for large-scale foundation model development.

Several directions remain for future investigations. This work evaluates CLN on a single dataset and two CNN-based architectures. Broader validation across tissue types and extension to transformer-based models \cite{chen2022scaling,xu2024vision}, including Vision Transformers and hierarchical attention architectures, is a natural and important next step. Furthermore, while pixel size serves as the sole conditioning signal, the CLN framework extends naturally to other axes of domain shift in computational pathology, including staining protocol, scanner vendor, and tissue preparation \cite{salehi2020pix2pix,debel2021residual}. A metadata-aware architecture stands for a promising direction toward truly generalizable WSI analysis systems. Moreover, we plan to explore different conditioning strategies, e.g., based on mixture‑of‑experts or meta‑learning.

In summary, we propose an efficient, lightweight and extensible mechanism for scale-invariant histopathological image analysis, replacing per-magnification model ensembles with a single magnification-conditioned network and offering a practical path toward unified multi-resolution learning in computational
pathology.

\begin{credits}
\subsubsection{\ackname} This project has received funding from the European Union's Horizon 2020 research and innovation programme under grant agreement No 857533 and from the International Research Agendas Programme of the Foundation for Polish Science No MAB PLUS/2019/13. The publication was created within the project of the Minister of Science and Higher Education "Support for the activity of Centers of Excellence established in Poland under Horizon 2020" on the basis of the contract number MEiN/2023/DIR/3796. We gratefully acknowledge Poland’s high-performance computing infrastructure PLGrid (HPC Centers: ACK Cyfronet AGH) for providing computer facilities and support within computational grant no PLG/2026/019392. This work was partially supported by the Excellence Initiative Research University program at the AGH University of Krakow.

\subsubsection{\discintname}
The authors have no competing interests to declare that are
relevant to the content of this article.

\end{credits}

\bibliographystyle{splncs04}
\bibliography{Paper-0009}

\end{document}